# *Research Note*
## Finding a Path is Harder than Finding a Tree

**Christopher Meek**  meek@microsoft.com
*Microsoft Research,*
*Redmond, WA 98052-6399 USA*

**Abstract**

I consider the problem of learning an optimal path graphical model from data and show the problem to be NP-hard for the maximum likelihood and minimum description length approaches and a Bayesian approach. This hardness result holds despite the fact that the problem is a restriction of the polynomially solvable problem of finding the optimal tree graphical model.

## 1. Introduction

The problem of learning graphical models has received much attention within the Artificial Intelligence community. Graphical models are used to represent and approximate joint distributions over sets of variables where the graphical structure of a graphical model represents the dependencies among the set of variables. The goal of learning a graphical model is to learn both the graphical structure and the parameters of the approximate joint distribution from data. In this note, I present a negative hardness result on learning optimal path graphical models.

Path graphical models are an interesting class of graphical models with respect to learning. This is due the fact that, in many situations, restricting attention to the class of path models is justified on the basis of physical constraints or temporal relationships among the variables. One example of this is the problem of identifying the relative positions of loci on a segment of DNA (e.g., Boehnke, Lange & Cox, 1991). In addition, one might be interested in obtaining a total order over a set of variables for other purposes such as visualization (e.g., Ma & Hellerstein, 1999).

The main positive results on the hardness of learning graphical models are for learning tree graphical models. These have been presented for maximum likelihood (ML) criterion (Edmonds, 1967; Chow & Liu, 1968) and adapted to a Bayesian criterion by Heckerman, Geiger, & Chickering (1995). Two NP-hardness results for learning graphical models have appeared in the literature. Those are the NP-hardness of finding the optimal Bayesian network structure with in-degree greater than or equal to two using a Bayesian optimality criterion (Chickering, 1996) and the problem of finding the ML optimal polytree (Dasgupta, 1999).

In this note, I present a proof of the hardness of finding an optimal path graphical models for the maximum likelihood (ML) criterion, the minimum description length (MDL) criterion, and a Bayesian scoring criterion. Unlike the ML hardness result of Dasgupta, I provide an explicit construction of a polynomial sized data set for the reduction and, unlike the Bayesian hardness result of Chickering (1996), I use a common "uninformative" prior.

## 2. Optimal Graphical Models

One of the primary goals when learning a graphical model is to obtain an approximate joint distribution over a set of variables from data. In this note, I focus on directed graphical models for a set of discrete variables $\{X_1, \ldots, X_n\}$. One component of a directed graphical model is its directed graphical structure that describes dependencies between the variables. A directed graphical model represents a family of distributions that factor according to the graphical structure $G$ of the directed graphical model, more specifically,

$$P_G(X_1, \ldots, X_n) = \prod_{i=1}^{n} P(X_i | pa_G(X_i))$$

where $pa_G(X_i)$ denotes the possibly empty set of parents of vertex $X_i$ in graph $G$. The subscript $G$ is omitted when it is clear from context. The most common methods guiding the choice of a distribution from a family of distributions are maximum likelihood estimation and Bayesian estimation. Given a graphical structure and a set of cases for the variables (also a prior distribution over the distributions in the case of the Bayesian approach), these methods provide an approximate joint distribution. For more details on graphical models and estimation see Heckerman (1998).

This leaves open the question of how one should choose the appropriate graphical structure. In the remainder of this section, I present the maximum likelihood (ML) criterion, the minimum discrimination length (MDL) criterion, and a Bayesian criterion for evaluating directed graphical models given a set of cases $D$. A value of the variable $X_i$ is denoted by $x_i$ and a value of the set of variables $pa(X_i)$ is denoted by $pa(x_i)$. The number of cases in $D$ in which $X_i = x_i$ and $pa(X_i) = pa(x_i)$ is denoted by $N(x_i, pa(x_i))$ and the total number of cases in $D$ is denoted by $N$.

One important property common to these scoring criteria is that the scores factor according to the graphical structure of the model. That is, the score for a graph $G$ and data set $D$ can be written as a sum of *local scores* for each of the variables

$$Score(G, D) = \sum_i LocalScore(X_i, pa(X_i)).$$

The local score for a variable $X_i$ is only a function of the counts for $X_i$ and $pa(X_i)$ in the data set $D$ and the number of possible assignments to the variables $X_i$ and $pa(X_i)$. Thus the structure of the graphical model determines which particular variables and counts are needed in the computation of the local score for a variable.

The log maximum likelihood scoring criterion for a graphical model is

$$Score_{ML}(G, D) = \sum_i LocalScore_{ML}(X_i, pa(X_i))$$

$$LocalScore_{ML}(X_i, pa(X_i)) = N \times H_D(X_i | pa(X_i)) \qquad (1)$$

where $H_D(X_i | pa(X_i))$ is the empirical conditional entropy of $X_i$ given its parents, and is equal to

$$- \sum_{X_i, pa(X_i)} \frac{N(x_i, pa(x_i))}{N} \log \frac{N(x_i, pa(x_i))}{N(pa(x_i))}.$$

One practical shortcoming of the ML score is that in comparing two models with graphical structure $G$ and $G'$ where $G$ contains a proper subset of the edges of $G'$ the ML score will never favor $G$. Thus, when using an ML score to choose among models without restricting the class of graphical structures, a fully connected structure is guaranteed to have a maximal score. This is problematic due to the potential for poor generalization error when using the resulting approximation. This problem is often called *overfitting*. When using this principle it is best to restrict the class of alternative structures under consideration in some suitable manner.

The minimum description length score can be viewed as a penalized version of the ML score

$$\begin{aligned} Score_{MDL}(G, D) &= Score_{ML}(G, D) - \frac{d \log N}{2} \\ &= \sum_i LocalScore_{MDL}(G, D) \end{aligned}$$

$$\begin{aligned} LocalScore_{MDL}(X_i, pa(X_i)) &= \\ LocalScore_{ML} &- \frac{\#(pa(X_i)) \times (\#(X_i) - 1) \times \log N}{2} \end{aligned} \quad (2)$$

where $d = \sum_i (\#(pa(X_i)) \times (\#(X_i) - 1))$ and $\#(Y)$ is used to denote the number of possible distinct assignments for a set of variables $Y$ and the number of assignments for the empty set of variables is $\#(\emptyset) = 1$. The penalty term leads to more parsimonious models, thus, alleviating the overfitting problem described above.

Finally, a Bayesian score requires a prior over the alternative models and, for each model, a prior over the distributions. A commonly used family of priors for directed graphical models is described by Cooper & Herskovits (1992). In their approach, one assumes a uniform prior on alternative graphs, $P(G) \propto 1$, and an "uninformative" prior over distributions. These assumptions lead to the following scoring function;

$$\begin{aligned} Score_{Bayes}(G, D) &= \log P(D|G) + \log P(G) \\ &\propto \sum_i LocalScore_{Bayes}(X_i, pa(X_i)) \end{aligned}$$

$$\begin{aligned} LocalScore_{Bayes}(X_i, pa(X_i)) &= \\ \log \prod_{pa(x_i)} &\frac{(\#(X_i) - 1)!}{(\#(X_i) - 1) + N(pa(x_i)))!} \prod_{x_i} N(x_i, pa(x_i))! \end{aligned} \quad (3)$$

Although not as apparent as in the MDL score, the Bayesian score also has a built-in tendency for parsimony that alleviates the problems of overfitting. The hardness results presented below can be extended to a variety of alternative types of priors including the BDe prior with an empty prior model (see Heckerman et al. 1995).

The problem of finding the optimal directed graphical model for a given class of structures $\mathcal{G}$ and data $D$ is the problem of finding the structure $G \in \mathcal{G}$ that maximizes $Score(G, D)$.

## 3. NP-Hardness of Finding Optimal Paths

In this section, I consider the problem of finding the optimal directed graphical model when the class of structures is restricted to be paths. A directed graphical structure is a *path* if there is one vertex with in-degree zero and all other vertices have in-degree one. I show that the problem of finding the optimal path directed graphical model is NP-hard for the commonly used scoring functions described Section 2. To demonstrate the hardness of finding optimal paths the problem needs to be formulated as a decision problem. The decision problem version of finding the optimal path directed graphical model is as follows

> The optimal path (OP) decision problem: Is there a path graphical model with score greater than or equal to $k$ for data set $D$?

In this section I prove the following theorem.

**Theorem 1** *The optimal path problem is NP-Hard for the maximum likelihood score, the minimum description length score and a Bayesian score.*

To prove this, I reduce the Hamiltonian Path (HP) decision problem to the OP decision problem.

> The Hamiltonian path (HP) decision problem: Is there a Hamiltonian path in an undirected graph $G$?

A Hamiltonian path for an undirected graph $G$ is a non-repeating sequence of vertices such that each vertex in $G$ occurs on the path and for each pair of adjacent vertices in the sequence there is an edge in $G$. Let the undirected graph $G = \langle V, E \rangle$ have vertex set $V = \{X_1, \ldots, X_n\}$ and edge set $E$.

The HP decision problem is NP-complete. Loosely speaking, this means that the HP decision problem is as computationally difficult as a variety of problems for which no known algorithm exists that runs in time that is a polynomial function of the size of the input. Theorem 1 indicates that the OP decision problem is at least as difficult as any NP-complete problem. For more information about the HP decision problem and NP-completeness see Garey & Johnson (1979).

I reduce the HP decision problem for $G$ to the OP decision problem by constructing a set of cases $D$ with the following properties;

$$\#(X_i) = \#(X_j) \tag{i}$$

$$LocalScore(X_i, \emptyset) = LocalScore(X_j, \emptyset) = \gamma \tag{ii}$$

$$LocalScore(X_i, \{X_j\}) \in \{\alpha, \beta\} \qquad \alpha < \beta \tag{iii}$$

$$LocalScore(X_j, \{X_i\}) = LocalScore(X_i, \{X_j\}) \tag{iv}$$

$$LocalScore(X_i, \{X_j\}) = \beta \quad \text{iff} \quad \{X_i, X_j\} \in E \tag{v}$$

For such a data set, the problem of the existence of a Hamiltonian path is equivalent to the existence of a path graphical model with score equal to $k = \gamma + (|V| - 1) \times \beta$ where $|V| = n$ is the number of vertices in the undirected graph $G$. Thus, to reduce the HP problem to the OP problem one needs to efficiently construct a polynomial sized data set with these properties. In other words, by such a construction, a general HP decision problem can be transformed into an OP decision problem. Because the size of the input to the OP problem is a polynomial function of the size of the input for the HP problem, if one can find an algorithm solve the OP problem in polynomial time then all NP-complete problems can be solved in polynomial time.

I construct a data set for graph $G$ assuming that each variable is ternary to satisfy condition (i). For each pair of vertices $X_i$ and $X_j$ ($i < j$) for which there is an edge in $G$, add the following 8 cases in which every variable $X_k$ ($k \neq i, j$) is zero.

| $X_1 \ldots X_{i-1}$ | $X_i$ | $X_{i+1} \ldots X_{j-1}$ | $X_j$ | $X_{j+1} \ldots X_n$ |
|---|---|---|---|---|
| $0 \ldots 0$ | 1 | $0 \ldots 0$ | 1 | $0 \ldots 0$ |
| $0 \ldots 0$ | 1 | $0 \ldots 0$ | 1 | $0 \ldots 0$ |
| $0 \ldots 0$ | 1 | $0 \ldots 0$ | 1 | $0 \ldots 0$ |
| $0 \ldots 0$ | 1 | $0 \ldots 0$ | 2 | $0 \ldots 0$ |
| $0 \ldots 0$ | 2 | $0 \ldots 0$ | 1 | $0 \ldots 0$ |
| $0 \ldots 0$ | 2 | $0 \ldots 0$ | 2 | $0 \ldots 0$ |
| $0 \ldots 0$ | 2 | $0 \ldots 0$ | 2 | $0 \ldots 0$ |
| $0 \ldots 0$ | 2 | $0 \ldots 0$ | 2 | $0 \ldots 0$ |

For each pair of vertices $X_i$ and $X_j$ ($i < j$) for which there is not an edge in $G$, add the following 8 cases.

| $X_1 \ldots X_{i-1}$ | $X_i$ | $X_{i+1} \ldots X_{j-1}$ | $X_j$ | $X_{j+1} \ldots X_n$ |
|---|---|---|---|---|
| $0 \ldots 0$ | 1 | $0 \ldots 0$ | 1 | $0 \ldots 0$ |
| $0 \ldots 0$ | 1 | $0 \ldots 0$ | 1 | $0 \ldots 0$ |
| $0 \ldots 0$ | 1 | $0 \ldots 0$ | 2 | $0 \ldots 0$ |
| $0 \ldots 0$ | 1 | $0 \ldots 0$ | 2 | $0 \ldots 0$ |
| $0 \ldots 0$ | 2 | $0 \ldots 0$ | 1 | $0 \ldots 0$ |
| $0 \ldots 0$ | 2 | $0 \ldots 0$ | 1 | $0 \ldots 0$ |
| $0 \ldots 0$ | 2 | $0 \ldots 0$ | 2 | $0 \ldots 0$ |
| $0 \ldots 0$ | 2 | $0 \ldots 0$ | 2 | $0 \ldots 0$ |

For a set of cases constructed as described above, the pairwise counts for a pair of variables $X_i$ and $X_j$ connected by an edge in $G$ are

|   |   | $X_i$ | | |
|---|---|---|---|---|
|   |   | 0 | 1 | 2 |
| $X_j$ | 0 | $4(n^2 - 5n + 6)$ | $4(n-2)$ | $4(n-2)$ |
|   | 1 | $4(n-2)$ | 3 | 1 |
|   | 2 | $4(n-2)$ | 1 | 3 |

The pairwise counts for a pair of variables $X_i$ and $X_j$ not connected by an edge in $G$ are

$$
\begin{array}{c|ccc}
 & \multicolumn{3}{c}{X_i} \\
 & 0 & 1 & 2 \\
\hline
X_j \quad 0 & 4(n^2 - 5n + 6) & 4(n-2) & 4(n-2) \\
1 & 4(n-2) & 2 & 2 \\
2 & 4(n-2) & 2 & 2 \\
\end{array}
$$

Condition (ii) is satisfied because the marginal counts for each variable are identical. There are two types of pairwise count tables, thus, there are at most two values for a given type of pairwise *LocalScore*. By using the two pairwise count tables and Equations 1, 2, and 3, one can easily verify that the local scores for the two tables satisfy condition (iii). It follows from the symmetry in the two types of pairwise tables and condition (ii) that condition (iv) is satisfied. It follows from the construction that condition (v) is satisfied. Furthermore, the set of cases is efficiently constructed and has a size which is polynomially bounded by the size of the graph $G$ proving the result.

## 4. Conclusion

In this note, I show that the problem of finding the optimal path graphical models is NP-hard for a variety of common learning approaches. The negative result for learning optimal path graphical models stands in contrast to the positive result on learning tree graphical models. This hardness result highlights one potential source of the hardness. That is, one can make an easy problem difficult by choosing an inappropriate subclass of models. Perhaps, by carefully choosing a broader class of models than tree graphical models one can identify interesting classes of graphical models for which the problem of finding an optimal model is tractable.

Another interesting class of graphical models not described in this note is the class of undirected graphical models (e.g., Lauritzen, 1996). The methods for learning undirected graphical models are closely related to the methods described in Section 2. In fact, for the case of undirected path models, the scoring formulas described in Section 2 are identical for each of the common approaches. Therefore, the NP-hardness result for directed path models presented in this note also applies to problem of learning undirected path models.

Finally, it is important to note that good heuristics exist for the problem of finding weighted Hamiltonian paths (Karp & Held, 1971). These heuristics can be used to identify good quality path models and rely on the fact that the optimal tree model can be easily found and will have a score at least as large as any path model.